\documentclass[11pt]{article}
\usepackage[margin=1in]{geometry}
\usepackage{amsmath,amssymb}
\usepackage{booktabs}
\usepackage{graphicx}
\usepackage{xcolor}
\usepackage{caption}
\usepackage[numbers,sort&compress]{natbib}
\usepackage{pgfplots}
\usepgfplotslibrary{groupplots}
\usetikzlibrary{calc}
\usepackage{url}
\pgfplotsset{compat=1.18}
\usepackage[colorlinks=true,linkcolor=blue!55!black,citecolor=blue!55!black,
            urlcolor=blue!55!black]{hyperref}
\hypersetup{
  pdftitle={Sample More, Reflect Less: Self-Refine and Reflexion Lose to
            Repeated Sampling at Equal Token Cost, from 1.5B to 7B},
  pdfauthor={Iliya Mirzaei},
  pdfsubject={Cost-matched evaluation of test-time reasoning methods for
              large language models},
  pdfkeywords={self-refine, reflexion, best-of-n, self-consistency,
               majority voting, self-correction, self-verification,
               test-time compute, inference scaling, compute-matched
               evaluation, chain-of-thought, multi-agent debate, GSM8K,
               MATH-500, large language models, LLM reasoning}
}

\definecolor{scblue}{HTML}{0072B2}
\definecolor{c1}{HTML}{D55E00}
\definecolor{c2}{HTML}{009E73}
\definecolor{c3}{HTML}{CC79A7}
\definecolor{c4}{HTML}{E69F00}
\definecolor{c5}{HTML}{56B4E9}
\definecolor{c6}{HTML}{6A3D9A}
\definecolor{c7}{HTML}{B15928}

\title{\bf Sample More, Reflect Less\\
\large Self-Refine and Reflexion Lose to Repeated Sampling\\
at Equal Token Cost, from 1.5B to 7B}

\author{Iliya Mirzaei\\
Department of Computer Science\\
Stony Brook University\\
\texttt{imirzaei@cs.stonybrook.edu}}

\date{\today}

\begin{document}
\maketitle

\begin{abstract}
Many methods promise to make a language model reason better without retraining
it. The model is told to plan before it solves, to criticise and rewrite its
own answer, to reflect on its mistakes, to pick the best of several attempts,
or to argue with copies of itself. Nearly all of these methods also make the
model write far more text than a single chain of thought, and writing more text
raises accuracy by itself. So a gain over one chain of thought does not show
that the method's idea is what helped.

\citet{wang2024tokeneconomies} made this point and reported that a simple
baseline often wins once the budgets are made comparable. That baseline is to
ask the same question several times and keep the answer that comes up most
often. Their evidence is a set of point estimates, with no confidence
intervals, no significance tests, and $100$ questions per dataset.

We run the comparison again as a designed experiment: seven methods, open
models of $1.5$B, $3$B and $7$B parameters, two mathematics benchmarks, $150$
questions each. We count every token a method generates, including the tokens
spent on critiques, reflections, debate turns and checking, and compare each
method against repeated sampling at that method's own measured cost. All $36$
comparisons are paired question by question and reported with confidence
intervals and a correction for testing many methods at once.

No method is reliably better than repeated sampling at equal cost in any
setting. Ten are reliably worse, all of them methods in which the model
inspects its own output, and two of those survive a correction applied across
all $36$ comparisons at once. Every one of the $18$ comparisons involving
self-inspection comes out negative.

The interesting part is what happens as the model grows, because the two kinds
of self-inspection part company. \emph{Choosing} among your own samples stops
being harmful: Best-of-$N$ draws eight samples and asks the model to pick the
best, and simply counting the most common answer instead is better by $8.0$ and
$11.3$ points at $1.5$B but only $2.0$ and $1.3$ points at $7$B, where the
difference is no longer distinguishable from zero. \emph{Rewriting} does not
recover: Self-Refine and a forced version of Reflexion remain significantly
below the equal-cost baseline at $7$B, by $3.6$ to $10.1$ points. Spending
tokens to reconsider an answer is a worse use of them than spending the same
tokens on another attempt, at every size we can test.

We also found that Reflexion, implemented as its authors describe it, never
once triggered its own retry on the smallest model. It judged itself correct on
every question, quietly turned into a single chain of thought, and scored well
because it had become cheap. A method that decides for itself when to act can
stop acting without any outward sign. We release the code, the prompts, every
generation, and the scripts we used to check our own numbers.

\end{abstract}

\section{Introduction}
\label{sec:intro}

A large family of methods claims to make language models reason better
without changing the model itself. The model is asked to think step by step
\citep{wei2022cot,kojima2022zeroshot}, to plan before it solves
\citep{wang2023plansolve}, to break the problem into smaller ones
\citep{zhou2023leasttomost}, to criticise and rewrite its own answer
\citep{madaan2023selfrefine}, to reflect on its mistakes and try again
\citep{shinn2023reflexion}, or to argue with copies of itself until they
agree \citep{du2024debate}.

These methods have something in common that is easy to overlook: almost all
of them make the model generate \emph{more text}. A single chain of thought
might be $300$ tokens. Three rounds of self-criticism might be $1{,}600$. A
debate between three copies of the model over two rounds might be $2{,}500$.
The method is not the only thing that changed. The compute budget changed
too, often by a factor of five or ten. And there is a much simpler way to
spend a larger budget: ask the model the same question several times and keep
the answer it gives most often \citep{wang2023selfconsistency}.

\paragraph{This problem is known.}
\citet{wang2024tokeneconomies} made exactly this argument and tested it.
Evaluating seven strategies across six datasets, they concluded that
``when we provide a simple baseline like chain-of-thought self-consistency
with comparable compute resources, it frequently outperforms reasoning
strategies proposed in the literature.'' We take that claim as our starting
point rather than our contribution.

\paragraph{But it has never been tested statistically.}
The evidence for that conclusion consists of point estimates. The study
reports no confidence intervals, no error bars, no variance across runs, and
no significance tests, and each dataset is represented by $100$ sampled
questions.

It is worth being concrete about what that sample size buys, using our own
measurements rather than an assumed variance. Across our $28$ comparisons at
$150$ questions, the standard error of a paired difference in accuracy runs
from $1.3$ to $3.2$ percentage points. Scaled to $n=100$ that is roughly
$1.6$ to $3.9$ points, so a $95\%$ interval would be about $\pm 3$ to
$\pm 8$ points wide. It is wider still if the comparison is not paired on
questions, which costs roughly a further factor of $\sqrt{2}$. Many of the
differences at stake are smaller than that. A conclusion of the form ``the
simple baseline frequently wins'' is therefore exactly the kind of claim that
can be produced by noise when several methods are compared without
correction, and whether it survives proper inference is unknown.

\paragraph{What we do.}
We re-run the comparison as a designed experiment. For seven methods, two
benchmarks, and two models, we measure accuracy and exact token cost, compare
each method against self-consistency \emph{at that method's own measured
cost}, and report every difference with a paired bootstrap confidence
interval and a $p$-value corrected for testing several methods at once. We also extend the question
to a regime the earlier study did not reach: models of $1.5$B and $3$B
parameters, well below its smallest ($7$B). This matters because the methods
under test rely on the model judging its own work, and small models are worse
at that. But repeated sampling also needs variety among the answers, which
small models may lack, so the direction of the effect is not obvious in
advance.

\paragraph{What we find.}
No method beats repeated sampling at equal cost in any of our four settings,
and six comparisons are significantly worse. But the useful result is not the
tally. Sorting the methods by \emph{what they buy with the extra tokens}
separates them cleanly: every method in which the model is asked to assess or
rewrite its own output falls below the cost-matched baseline in all twelve of
its comparisons, while methods that add no self-assessment sit on the
baseline.

One method lets us test that explanation without any confound at all.
Best-of-$N$ draws eight samples and then asks the model to pick the best one.
We can take those \emph{same eight samples} and simply count which answer
occurs most often. The samples, the tokens, and the model are identical; only
the final step differs. Counting wins in every setting at these sizes, by
between $5$ and $17$ percentage points.

Because that comparison is cheap, we could also ask where it stops holding. We
ran it on a $7$B model as well, and there the penalty falls to about two points
and is no longer distinguishable from zero. The claim we end up with is
therefore a bounded one: choosing is worse than counting below $7$B, and the
two draw level at $7$B.

\paragraph{Contributions.}
\begin{enumerate}
\itemsep2pt
\item A controlled demonstration that self-evaluation, not extra computation,
      is where these methods lose: with the sampled candidates held fixed, a
      model-chosen answer is worse than a counted one
      (Section~\ref{sec:results}).
\item A statistical re-examination of the budget-matched comparison: paired
      bootstrap intervals, Holm correction across methods, and an explicit
      statement of the effect size the design can detect
      (Sections~\ref{sec:setup} and \ref{sec:results}).
\item A finer budget-matching procedure. Rather than fixing a common query or
      token cap for all methods, we measure each method's actual cost and
      compare it against the baseline curve interpolated at that cost
      (Section~\ref{sec:idea}).
\item An efficient construction of the entire self-consistency
      accuracy-versus-cost curve from a single pool of samples, which is what
      makes per-method matching affordable (Section~\ref{sec:idea}).
\item Evidence from the small-model regime ($1.5$B and $3$B), which the prior
      study does not cover.
\item A complete open release: harness, prompts, and every raw generation, so
      that answer extraction and grading can be re-checked rather than
      trusted. The prior study released none of these.
\end{enumerate}

\paragraph{Scope.}
This is a replication with extensions, not a new phenomenon. Our contribution
is to establish how much of the earlier conclusion is supported once
uncertainty is quantified, and whether it holds for small open models. We
study mathematical reasoning with automatically checkable answers on CPU
hardware; we do not test frontier models, and we say plainly in
Section~\ref{sec:discussion} what that does and does not license.

\section{The idea: compare at equal cost}
\label{sec:idea}

\subsection{Why the usual comparison is unfair}

Take Self-Refine as an example. The model writes an answer, then criticises
its own answer, then rewrites it. With three rounds of criticism that is
seven calls to the model instead of one. If the rewritten answer is better
than the first answer, we have learned something, but not necessarily
that criticising helps. We have spent seven times the compute, and spending
more compute is known to help on its own.

The fair question is: \emph{if we had simply spent those seven calls on
seven independent attempts and taken the most common answer, would we have
done better or worse?} That is the comparison this paper makes.

\subsection{The baseline: just sample more}

Our reference point is self-consistency \citep{wang2023selfconsistency}. It
is the simplest possible way to convert extra compute into extra accuracy:
draw $N$ independent chains of thought at a non-zero temperature, read the
final answer off each one, and return whichever answer appears most often.
It involves no critique, no planning, no communication between attempts, and
no extra prompt engineering.

Because self-consistency can be run at any $N$, it does not give a single
number. It gives a \emph{curve} of accuracy against cost. That curve is
what we compare against. We call it the \emph{sampling baseline}. A method is
worth its cost only if it lands \emph{above} this curve.

\subsection{Measuring cost}

We measure cost as the number of tokens the model generates, summed over
every call a method makes for a single question. This includes tokens that
never appear in the final answer: critiques, self-evaluations, reflections,
messages exchanged between debating agents, and verification steps. Those
tokens are generated, so they are paid for.

We count generated (completion) tokens rather than wall-clock time because
time depends on the hardware, the batch size, and how much of the work can
be run in parallel, none of which are properties of the method itself.
Section~\ref{sec:discussion} discusses what changes if one counts input
tokens or latency instead.

\subsection{Getting the whole baseline curve cheaply}

Running self-consistency separately at each $N$ we care about would mean
generating a fresh batch of chains for every one of them. Instead we generate
a \emph{pool} of $K = 16$ independent chains once per question, and then read
off self-consistency at any $N \le K$ by repeatedly drawing $N$ chains from
the pool without replacement and taking the majority vote. We evaluate at
$N \in \{1, 2, 3, 4, 6, 8, 12, 16\}$, averaging over $200$ draws at each
value. Each draw yields both an outcome (was the majority answer correct?)
and a cost, the summed length of exactly those $N$ chains, and we read the
two off the \emph{same} draw, so every point on the curve describes a run
that could actually have happened.

This is not an approximation of a different experiment: drawing $N$ chains
from a pool of $K$ independent chains is distributed exactly as running
self-consistency at $N$ would be. Reusing the pool costs us only Monte-Carlo
error from using $200$ draws instead of all $\binom{K}{N}$ of them, which we
quantify in Section~\ref{sec:results}.

\section{Related work}
\label{sec:related}

\paragraph{Methods that spend more at test time.}
Chain-of-thought prompting showed that asking a model to write out
intermediate steps improves reasoning \citep{wei2022cot}, and that a single
instruction is often enough to trigger it \citep{kojima2022zeroshot}.
Self-consistency samples several chains and takes a majority vote
\citep{wang2023selfconsistency}. Plan-and-Solve asks the model to write a
plan first \citep{wang2023plansolve}, and least-to-most decomposes a problem
into easier sub-problems \citep{zhou2023leasttomost}. Self-Refine has the
model critique and rewrite its own output \citep{madaan2023selfrefine}, and
Reflexion adds a memory of past failures \citep{shinn2023reflexion}. Tree of
Thoughts searches over partial solutions \citep{yao2023tot}, Graph of Thoughts
generalises that search to a graph \citep{besta2024got}, progressive-hint
prompting feeds previous answers back as hints \citep{zheng2023progressive},
and multi-agent debate runs several copies of the model against each other
\citep{du2024debate}. All of these increase the number of tokens generated
per question, by factors ranging from roughly two to well over ten.

\paragraph{Doubts about self-correction.}
The closest prior result to ours concerns self-correction specifically.
\citet{kamoi2024correct} survey the area and find that reported gains often
depend on conditions that do not hold in practice, such as access to reliable
external feedback. \citet{huang2024selfcorrect} showed that when a model is not told whether its
answer was right, asking it to correct itself often makes things
\emph{worse}, and that reported gains frequently come from using the correct
answer to decide when to stop. Our study is broader in scope, covering
planning, debate, and verification as well as self-correction, and it differs
in what it controls for: \citeauthor{huang2024selfcorrect} control the
\emph{information} available to the method, whereas we control the
\emph{budget} it consumes. The two controls are complementary, and we adopt
theirs as well by never revealing ground truth to any method.

\paragraph{Budget-aware evaluation: the work we replicate.}
\citet{wang2024tokeneconomies} is the direct predecessor of this paper and
the source of the claim we test. They evaluate seven strategies
(chain-of-thought self-consistency, multi-agent debate, Reflexion,
Plan-and-Solve, least-to-most, progressive-hint prompting, and Tree of
Thoughts) on six datasets including GSM8K and MATH, using GPT-3.5, GPT-4,
Mistral-7B, LLaMA-2-70B, and Mixtral-8x7B. They define budget as input plus
output tokens and impose a shared cap of $20$ queries or $10$k tokens, and
report that self-consistency ``consistently beat other reasoning strategies
across all 5 datasets with significantly less budget.''

We differ in four ways, none of which concern the idea of matching budgets,
which is theirs. First, they report point estimates only, with no confidence
intervals, error bars, run-to-run variance or significance tests, on
$100$ questions per dataset, which is not enough to separate differences
below roughly ten percentage points from noise; we quantify the uncertainty.
Second, their budget matching is a shared cap applied to every method, so
methods are compared at whatever cost the cap happens to induce; we measure
each method's realised cost and compare it against the baseline curve
interpolated at exactly that cost. Third, their smallest model has $7$B
parameters, leaving the small-model regime untested. Fourth, they release
neither code nor generations, so their grading cannot be audited; we release
both.

\paragraph{What is already known about verifiers, and what is not.}
Our sharpest result concerns Best-of-$N$, so we should be careful about how
much of it is new. It is established that selecting among samples with an
imperfect verifier has limits. \citet{stroebl2024flaws} show that resampling
only keeps paying off if the verifier is perfect, because an imperfect one
lets false positives through, and that this bounds how far Best-of-$N$ can go.
It is also reported that selection with external or trained reward models
often fails to beat plain majority voting. Verifiers that are trained for the
job are a different matter: process supervision \citep{lightman2024verify} and
verifiers trained alongside the generator \citep{hosseini2024vstar} both help,
and \citet{zhang2025selfverify} find that a trained self-verifier beats
majority voting at $7$B. Cheaper selection signals that avoid a second model,
such as the generator's own confidence \citep{kang2025selfcertainty}, sit
between these cases.

Our case is the one in between, and it is the one people actually deploy: the
same model, with no training and no reward model, asked in a single prompt to
pick the best of its own samples. We are not aware of a published measurement
of that specific comparison against majority voting on identical samples, and
we provide one. But we want to be plain that the direction is what the
verifier literature would predict. The contribution is the size of the gap,
the fact that it holds in every setting we test, and the fact that it is
measured with the samples held fixed so that nothing else can explain it.
Taken with \citet{zhang2025selfverify}, the natural reading is that the
training, not the verifying, is what makes a verifier useful.

\paragraph{Related negative results on debate.}
\citet{choi2025debatevote} show theoretically and empirically that
simultaneous-update debate forms a martingale on agents' belief in the
correct answer, implying no expected gain beyond what majority voting already
provides, and report that majority voting accounts for most of the measured
benefit of multi-agent debate. \citet{tran2026singleagent} reach a compatible
conclusion for multi-hop question answering under matched ``thinking token''
budgets. Our debate results should be read as an independent check of these
findings in a different regime rather than as a new discovery.

\paragraph{Cost-matched comparison elsewhere.}
Outside of audits, comparing at equal compute appears mostly \emph{inside}
papers proposing a new method, as a way of showing that the new method beats
self-consistency at the same budget \citep{sharma2025sequential}.
A related line treats the budget as the object of study rather than as a
control: \citet{snell2024testtime} ask how best to allocate a fixed test-time
budget, \citet{wu2024inferencescaling} fit scaling laws for compute-optimal
inference, \citet{brown2024monkeys} show how far accuracy rises with repeated
sampling alone, and \citet{muennighoff2025s1} obtain strong results from a
deliberately simple budget-forcing scheme. Our baseline is the plainest member
of that family, used here as a control rather than as a proposal.

\paragraph{Systematic evaluations of prompting.}
\citet{preet2026simplicity} evaluate eight prompting techniques across ten
multiple-choice benchmarks and report that plain baseline prompting matches
or beats more elaborate techniques in most configurations. Their study and
ours point in a similar direction from different angles: they vary the
\emph{wording} of a single-shot prompt on multiple-choice questions, while we
vary the \emph{amount of computation} for multi-step and multi-sample methods
on open-ended answers. Neither subsumes the other.

\paragraph{Statistics in evaluation.}
\citet{miller2024errorbars} argue that language-model evaluations are
experiments and should be reported with error bars and significance tests,
which is often not done. We follow that advice: every number we report has a
confidence interval, all comparisons are paired on questions, and we correct
for testing several methods at once.

\section{Experimental setup}
\label{sec:setup}

\subsection{Models}

We use two instruction-tuned open-weight models from the Qwen2.5 family
\citep{qwen25}: Qwen2.5-1.5B-Instruct and Qwen2.5-3B-Instruct. Both are run
with \texttt{llama.cpp} \citep{llamacpp} at \texttt{Q8\_0} quantisation
(8 bits per weight) on CPU. All generations come from a single machine with
one AMD EPYC 7452 (32 cores, 64 threads) and 125\,GB of RAM; there is no GPU.
Section~\ref{sec:discussion} discusses what quantisation and model scale
mean for how far these results generalise.

\subsection{Benchmarks}

We use two mathematical reasoning benchmarks whose answers can be graded
automatically and exactly:

\begin{itemize}
\itemsep2pt
\item \textbf{GSM8K} \citep{cobbe2021gsm8k}: grade-school word problems with
      a single integer answer. We draw $150$ questions from the $1{,}319$
      test items.
\item \textbf{MATH-500} \citep{hendrycks2021math}: a $500$-problem subset of
      competition mathematics, with answers written as formulas such as
      $\left(3,\tfrac{\pi}{2}\right)$. We draw $150$ of them.
\end{itemize}

This is half again as many questions per dataset as the $100$ used by
\citet{wang2024tokeneconomies}, which matters because the size of that sample
is one of the reasons their conclusion could not be tested statistically.

Both samples are drawn once with a fixed random seed and then held constant
across every method and model, so all comparisons are on identical questions.
We deliberately avoid multiple-choice benchmarks, where a method can score
above chance without producing a correct derivation.

\subsection{Methods compared}

Table~\ref{tab:methods} lists the seven methods and the number of model calls
each makes. We implement each method from the description in its original
paper, with one uniform restriction: \emph{no method is ever shown the
correct answer}. This matters for the self-correcting methods, where using
ground truth to decide when to stop would leak the answer into the procedure
\citep{huang2024selfcorrect}.

\begin{table}[t]
\centering
\small
\begin{tabular}{llc}
\toprule
Method & Configuration & Model calls per question\\
\midrule
Chain-of-Thought \citep{wei2022cot} & greedy, one sample & $1$\\
Plan-and-Solve \citep{wang2023plansolve} & greedy, one sample & $1$\\
Self-Refine \citep{madaan2023selfrefine} & $R=3$ critique/revise rounds & $7$\\
Reflexion \citep{shinn2023reflexion} & $R=3$ rounds, early stop & $1$ to $10$\\
Reflexion (forced) & $R=3$ rounds, no early stop & $7$\\
Best-of-$N$ + self-verify & $N=8$, model picks & $9$\\
Multi-Agent Debate \citep{du2024debate} & $A=3$ agents, $R=2$ rounds & $9$\\
\midrule
\emph{Sampling baseline} \citep{wang2023selfconsistency}
  & pool of $K=16$, subsampled & $1$ to $16$\\
\bottomrule
\end{tabular}
\caption{The methods under test. Reflexion's cost varies because it stops
early when the model judges its own answer to be correct.}
\label{tab:methods}
\end{table}

\paragraph{Why there are two versions of Reflexion.}
Reflexion as specified decides for itself when to stop: after each attempt the
model is asked whether its own answer is correct, and it reflects and retries
only if it says no. How often that happens turned out to depend heavily on the
model. On Qwen2.5-1.5B the answer was ``correct'' on \emph{every single
question} in both benchmarks, so the loop always exited immediately and the
method silently collapsed into one chain of thought. On Qwen2.5-3B it fires
more often, and on MATH-500 it fires most of the time; the exact rates are in
Section~\ref{sec:results}.

Measuring only this version would therefore tell us little about whether
reflection helps, because on the smaller model reflection never happened at
all. We also run a \emph{forced} variant that skips the self-assessment and
always performs its three reflect-and-retry rounds.

We want to be exact about what each version licenses, because it would be easy
to criticise a method we had altered. The first version is Reflexion as
\citet{shinn2023reflexion} describe it, and it is the only one of the two that
supports any claim about their method. The forced variant is \emph{ours}, not
theirs. It exists to answer a different question: given that the published
procedure did nothing on the smaller model, does the reflect-and-retry
mechanism help when it is made to run? Statements about ``Reflexion'' in this
paper refer to the published version. Statements about reflection as a
mechanism refer to the forced one, and we label it as such everywhere it
appears.

Deterministic methods (Chain-of-Thought, Plan-and-Solve, Self-Refine, and
both versions of Reflexion) are run at temperature $0$. Methods that rely on diversity between
samples (the sampling baseline, Best-of-$N$, Debate) are run at temperature
$0.7$. Generation is capped at $1{,}024$ tokens per call for every method
equally. Each question--method pair gets a seed derived deterministically
from the model, dataset, method, configuration, and question index, so the
entire study can be reproduced exactly.

\subsection{Grading}

Answers are extracted by a deterministic parser: the content of the last
\verb|\boxed{...}| if present, otherwise the text following ``the answer
is'', otherwise the last number in the response. Extracting the braced
content requires real brace matching rather than a regular expression,
because answers such as $\frac{3\sqrt{3}}{4}$ nest braces to arbitrary depth
and a fixed-depth pattern silently returns nothing. For GSM8K the extracted
value is compared numerically with a tolerance of $10^{-6}$. For MATH-500 we
normalise common LaTeX variants (\verb|\left|/\verb|\right|, \verb|\dfrac|
versus \verb|\frac|, degree symbols, surrounding braces) before comparing as
strings, and additionally accept a numeric match.

\paragraph{Validating the grader.}
A grading bug depresses every method equally and is therefore invisible in a
comparison, so we test the grader directly. We construct, for each of the
$1{,}319$ GSM8K and $500$ MATH-500 test items, a synthetic response that
states the gold answer inside \verb|\boxed{...}|, under six cosmetic
variations: verbatim, extra spacing before control sequences, \verb|\dfrac|
substituted for \verb|\frac|, parentheses wrapped in
\verb|\left|/\verb|\right|, a trailing period, and an extra enclosing brace
pair. The parser recovers and grades the answer correctly in
$500/500$ MATH-500 items and $1{,}319/1{,}319$ GSM8K items under
\emph{all six} variations. Any residual failures in the real experiment are
therefore attributable to the model, not to the grader.

The identical grading code is used by the experiment runner and by the
analysis. We store the raw text of every generation and re-grade from that
text at analysis time, so runner-time and analysis-time grading cannot
diverge, and so a reader who disagrees with our parser can substitute their
own without regenerating anything. We report the observed rate of
unparseable model answers in Section~\ref{sec:results}.

\subsection{Statistics}

Every comparison is paired: method and baseline are evaluated on the same
$150$ questions, so we compare per-question outcomes rather than two
independent accuracy numbers.

For a method with measured mean cost $c$, we locate the point on the sampling
baseline curve with the same cost. Because the baseline is only defined at
integer $N$, we linearly interpolate between the two values of $N$ whose mean
costs bracket $c$, in (cost, accuracy) space.

Confidence intervals come from a paired bootstrap \citep{efron1979bootstrap}
over questions with $10{,}000$ resamples: questions are the unit of resampling, since they are
what we wish to generalise over. We report the mean difference in accuracy,
its $95\%$ interval, and a two-sided $p$-value. Testing seven methods at once
makes it likely that one of them looks unusual by chance, so within each
setting the $p$-values are adjusted using the Holm--Bonferroni procedure
\citep{holm1979}.

\subsection{What we can and cannot detect}

This is the question the prior study did not ask, so we state the answer
plainly. With $150$ paired questions, the standard error of a paired
difference in accuracy across our $28$ comparisons runs from $1.3$ to $3.2$
percentage points, with a median of $2.6$; the corresponding $95\%$ intervals
are between $\pm 2.6$ and $\pm 6.3$ points wide. The design can therefore
identify differences of roughly five to six percentage points or larger in the
typical case; smaller true differences cannot be reliably separated from
zero.

Two consequences follow, and we hold to both in
Section~\ref{sec:results}. First, a result of ``no significant difference''
means the data are consistent with anything inside the interval. It is not
evidence that the true effect is zero. We therefore always report the
interval, and describe what magnitude of benefit it still permits. Second,
because seven methods are compared in each setting, some difference will look
large by chance; the Holm correction is what keeps that from being read as a
finding.

\subsection{Threats to validity}
\label{sec:threats}

We list the design choices most likely to affect the conclusions, and what we
did about each.

\paragraph{Temperature is confounded with method.}
The deterministic methods run at temperature $0$, while the sampling baseline
requires a non-zero temperature to produce diverse samples. A difference
between them could therefore reflect temperature rather than method. Our
design contains the control for this: self-consistency at $N=1$ is a single
sample at temperature $0.7$ and costs almost exactly what one greedy chain of
thought costs. Comparing greedy chain-of-thought against $\text{SC}@1$
isolates the temperature effect from everything else, and we report it
separately in Section~\ref{sec:results}.

\paragraph{Interpolating the baseline curve.}
Self-consistency is defined only at integer $N$, so matching a method's cost
exactly requires interpolating between two adjacent values of $N$. We
interpolate linearly in (cost, accuracy) space. Over this range the baseline's accuracy rises quickly at first and then
flattens, so a straight line drawn between two points on it sits \emph{below}
the true curve. That makes the baseline slightly weaker than it really is. This biases our comparison in favour of
the methods under test, not against them.

\paragraph{We measure question variance, not run-to-run variance.}
This is the most important limitation of our statistics and we want it stated
before the results rather than after. Our bootstrap resamples \emph{questions},
so every interval answers ``how much would this differ on another sample of
questions from the same benchmark''. It does not answer ``how much would this
differ if the same questions were run again with different random seeds''.
Three of our methods involve randomness: the sampling baseline, Best-of-$N$
and debate. Each was executed once, so their seed-level variance is
absorbed silently rather than estimated. The deterministic methods are run at
temperature $0$ and repeat exactly, so for them the question is moot. Since
\citet{miller2024errorbars}, whose advice we otherwise follow, recommends
accounting for both sources, our intervals should be read as lower bounds on
total uncertainty for the three methods that involve randomness. The direction of that bias is
towards \emph{over}-stating significance, so the six significant results
should be treated as the least secure part of our findings. The Best-of-$N$
comparison in Section~\ref{sec:results} is not affected, because there the
samples are held fixed and only the selection rule changes.

\paragraph{Cost matching at the cheap end.}
Greedy chain-of-thought turns out to cost marginally less than a single
sampled chain, $292$ against $297$ tokens on one setting and similarly
elsewhere, because greedy decoding produces slightly shorter solutions. The
baseline curve does not extend below $N=1$, so in three of the four settings
chain-of-thought is compared against $\text{SC}@1$ rather than against a
cheaper interpolated point. The mismatch is at most five tokens out of roughly
three hundred and cannot affect any conclusion, but it means the comparison
for that one method is against a baseline costing very slightly more. At the
expensive end no such issue arises: no method costs more than
$\text{SC}@16$ in any setting, so nothing is extrapolated beyond the measured
curve.

\paragraph{Reflexion's budget is chosen by the method.}
Reflexion decides for itself how many rounds to run, so the cost we match it
at is a cost it selected rather than one we imposed. This is the right
comparison for a method one would actually deploy, but it does mean that
Reflexion's position on the cost axis is an outcome of the experiment and not
a design parameter. The forced variant, whose budget is fixed in advance,
does not have this property.

\paragraph{Monte Carlo error from subsampling.}
Each point on the baseline curve is estimated by drawing $200$ subsets from
the pool rather than enumerating all $\binom{K}{N}$ of them. This adds a
small amount of noise that our question-level bootstrap does not capture. We
quantify its size in Section~\ref{sec:results} by re-running the analysis
with a different random seed and reporting how much the estimates move.

\paragraph{How wide we cast the correction.}
Testing seven methods at once makes it likely that one looks unusual by
chance, so we apply the Holm correction across the seven methods within each
model and dataset. That controls the chance of \emph{any} false alarm inside
a setting. A stricter reading would correct across all four settings at once.
We report both, so that neither choice has to be taken on trust.

\paragraph{Is the verifier's prompt doing the damage?}
Best-of-$N$ carries our sharpest claim, so the obvious objection is that we
simply asked the model to judge badly. Three things limit that reading, and
one does not.

First, the verifier is not being asked to do anything subtle. It sees the
problem and eight complete solutions and is asked which is most likely
correct. Second, it is not answering at random: it agrees with the majority
answer on $63$ to $89$ per cent of questions, so it is tracking something
real, and it loses ground specifically on the questions where it departs from
the majority. Third, on the smaller model it picks the first candidate about
three-quarters of the time, which is a position bias rather than a reasoning
failure. A preference for whichever option is presented first is a documented
property of language models asked to choose between candidates
\citep{zheng2023judging}, not an artefact of our wording.

What we cannot rule out is that a longer or more structured rubric, or a
verifier prompted to score each candidate separately rather than choose among
them, would do better. Our claim is therefore about the cheap self-check that
Best-of-$N$ as usually described implies, not about every possible way of
asking a model to evaluate. A reader who wants to test a better rubric can do
so against our stored candidates without regenerating them.

\paragraph{Where does debate belong?}
We describe debate as ending in a count rather than a judgement, because its
final step is a majority vote over the agents. That is not the whole story:
each agent also reads and revises against the others, which is a form of
mutual assessment. Debate therefore sits between our two groups, and it
behaves that way, losing less to the baseline than the self-assessing methods
but more than the single-pass ones. We flag this because a clean two-way split
would be tidier than the evidence warrants.

\paragraph{Prompt wording, and Reflexion in particular.}
Each method is one implementation of a description in prose, and prompt
wording matters. This bears hardest on our finding that Reflexion's
self-assessment never fires on the smaller model. That behaviour is a property
of the model's response to \emph{our} judging prompt, not a theorem about
Reflexion: a differently worded question might elicit ``incorrect'' more
often. We report the exact wording in Appendix~\ref{app:repro} so the claim can
be checked rather than believed, and the forced variant exists precisely so
that a conclusion about the reflection mechanism does not rest on the
judging prompt at all.

\paragraph{One configuration per method.}
Each method has settings such as rounds, agents and samples that we
fixed in advance rather than tuning. Tuning them per dataset would raise
every method's accuracy, but choosing the best configuration requires the
correct answers, which is the leak that \citet{huang2024selfcorrect}
identify. Fixing configurations in advance avoids that leak at the cost of
possibly under-representing each method's best case. The same fixed
configuration is used for both models and both datasets.

\paragraph{Cost measure, and what happens if we charge for input.}
We recorded generated tokens only. That was a deliberate choice, because
generated tokens are the part of the cost that does not depend on how a
deployment caches or batches prompts, but it is also the accounting under
which the methods we test look \emph{best}, so it deserves a direct answer
rather than an argument.

For the two methods in our central comparison the input side is not lost. It
is fully determined by things we did store: the sampling baseline sends the
same question prompt once per sample, and Best-of-$N$ sends that prompt $N$
times plus one verification prompt containing all $N$ candidate solutions,
which we stored verbatim. We therefore reconstructed both prompts and counted
them with the model's own tokenizer.

The asymmetry is large. Best-of-$N$ reads more than it writes, at $1.25$ to
$1.42$ input tokens per output token, because the verification step re-reads
eight full solutions. The baseline reads far less, at $0.22$ to $0.37$, because
it only ever sends the question. Redoing the whole comparison in total-token
space, so that the baseline curve is interpolated at Best-of-$N$'s realised
total cost, moves the verdict against Best-of-$N$ in every setting: from
$-7.7$ to $-9.1$ points on Qwen2.5-1.5B with GSM8K, from $-8.1$ to $-9.3$ on
the same model with MATH-500, from $-14.1$ to $-16.8$ on Qwen2.5-3B with
MATH-500, and by smaller margins elsewhere. At $7$B the difference remains
statistically indistinguishable from zero under either measure.

We did not reconstruct input costs for Self-Refine, Reflexion or debate,
because those methods re-send intermediate critiques and agent messages that we
did not store. The direction for them is fixed by construction, since all three
re-read material the baseline never sends, but we do not put numbers on it and
do not rely on it anywhere.

\section{Results}
\label{sec:results}

We report 6 settings: Qwen2.5-1.5B on GSM8K, Qwen2.5-1.5B on MATH-500, Qwen2.5-3B on GSM8K, Qwen2.5-3B on MATH-500, Qwen2.5-7B on GSM8K, Qwen2.5-7B on MATH-500. Each uses 150 paired questions.

\subsection{Does any method beat the sampling baseline at equal cost?}

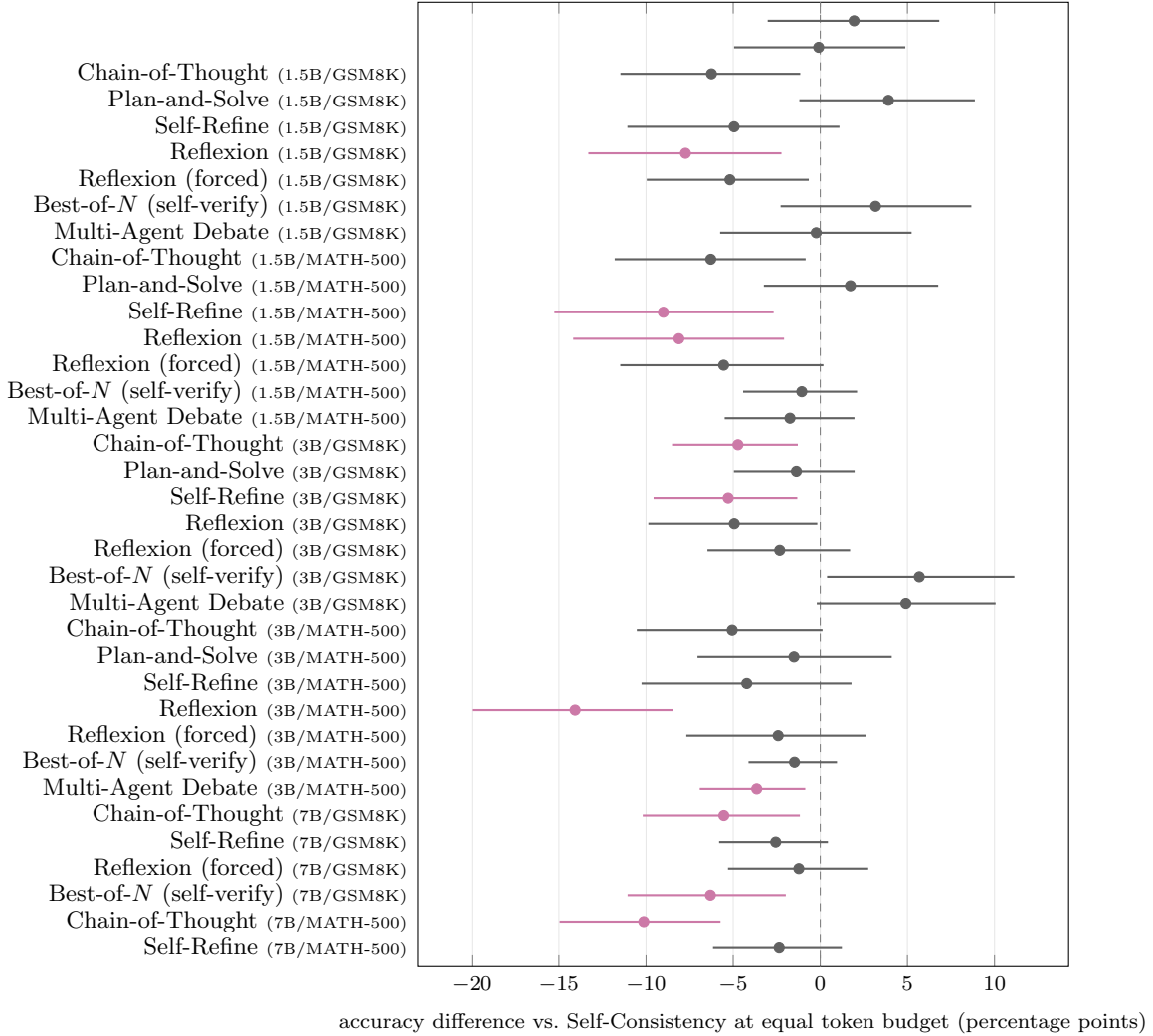
\begin{figure}[htbp]\centering
\begin{tikzpicture}
\begin{axis}[width=0.62\textwidth, height=14.4cm,
  xlabel={accuracy difference vs.\ Self-Consistency at equal token budget (percentage points)},
  ytick=data, yticklabels={{\footnotesize Best-of-$N$ (self-verify) \tiny(7B/MATH-500)},{\footnotesize Reflexion (forced) \tiny(7B/MATH-500)},{\footnotesize Self-Refine \tiny(7B/MATH-500)},{\footnotesize Chain-of-Thought \tiny(7B/MATH-500)},{\footnotesize Best-of-$N$ (self-verify) \tiny(7B/GSM8K)},{\footnotesize Reflexion (forced) \tiny(7B/GSM8K)},{\footnotesize Self-Refine \tiny(7B/GSM8K)},{\footnotesize Chain-of-Thought \tiny(7B/GSM8K)},{\footnotesize Multi-Agent Debate \tiny(3B/MATH-500)},{\footnotesize Best-of-$N$ (self-verify) \tiny(3B/MATH-500)},{\footnotesize Reflexion (forced) \tiny(3B/MATH-500)},{\footnotesize Reflexion \tiny(3B/MATH-500)},{\footnotesize Self-Refine \tiny(3B/MATH-500)},{\footnotesize Plan-and-Solve \tiny(3B/MATH-500)},{\footnotesize Chain-of-Thought \tiny(3B/MATH-500)},{\footnotesize Multi-Agent Debate \tiny(3B/GSM8K)},{\footnotesize Best-of-$N$ (self-verify) \tiny(3B/GSM8K)},{\footnotesize Reflexion (forced) \tiny(3B/GSM8K)},{\footnotesize Reflexion \tiny(3B/GSM8K)},{\footnotesize Self-Refine \tiny(3B/GSM8K)},{\footnotesize Plan-and-Solve \tiny(3B/GSM8K)},{\footnotesize Chain-of-Thought \tiny(3B/GSM8K)},{\footnotesize Multi-Agent Debate \tiny(1.5B/MATH-500)},{\footnotesize Best-of-$N$ (self-verify) \tiny(1.5B/MATH-500)},{\footnotesize Reflexion (forced) \tiny(1.5B/MATH-500)},{\footnotesize Reflexion \tiny(1.5B/MATH-500)},{\footnotesize Self-Refine \tiny(1.5B/MATH-500)},{\footnotesize Plan-and-Solve \tiny(1.5B/MATH-500)},{\footnotesize Chain-of-Thought \tiny(1.5B/MATH-500)},{\footnotesize Multi-Agent Debate \tiny(1.5B/GSM8K)},{\footnotesize Best-of-$N$ (self-verify) \tiny(1.5B/GSM8K)},{\footnotesize Reflexion (forced) \tiny(1.5B/GSM8K)},{\footnotesize Reflexion \tiny(1.5B/GSM8K)},{\footnotesize Self-Refine \tiny(1.5B/GSM8K)},{\footnotesize Plan-and-Solve \tiny(1.5B/GSM8K)},{\footnotesize Chain-of-Thought \tiny(1.5B/GSM8K)}},
  ytick style={draw=none}, y tick label style={font=\scriptsize, align=right},
  xmajorgrids, grid style={gray!18}, tick label style={font=\scriptsize},
  label style={font=\scriptsize},
  ymin=0.3, ymax=36.7]
\addplot[gray, dashed, forget plot, samples=2, domain=0:0] coordinates {(0,0) (0,37)};
\addplot[black!62, thick, forget plot] coordinates {(-6.16,1) (1.24,1)};
\addplot[only marks, mark=*, mark size=1.9pt, black!62, forget plot] coordinates {(-2.36,1)};
\addplot[c3, thick, forget plot] coordinates {(-14.96,2) (-5.73,2)};
\addplot[only marks, mark=*, mark size=1.9pt, c3, forget plot] coordinates {(-10.13,2)};
\addplot[c3, thick, forget plot] coordinates {(-11.06,3) (-1.98,3)};
\addplot[only marks, mark=*, mark size=1.9pt, c3, forget plot] coordinates {(-6.31,3)};
\addplot[black!62, thick, forget plot] coordinates {(-5.30,4) (2.75,4)};
\addplot[only marks, mark=*, mark size=1.9pt, black!62, forget plot] coordinates {(-1.23,4)};
\addplot[black!62, thick, forget plot] coordinates {(-5.81,5) (0.44,5)};
\addplot[only marks, mark=*, mark size=1.9pt, black!62, forget plot] coordinates {(-2.56,5)};
\addplot[c3, thick, forget plot] coordinates {(-10.19,6) (-1.17,6)};
\addplot[only marks, mark=*, mark size=1.9pt, c3, forget plot] coordinates {(-5.54,6)};
\addplot[c3, thick, forget plot] coordinates {(-6.92,7) (-0.86,7)};
\addplot[only marks, mark=*, mark size=1.9pt, c3, forget plot] coordinates {(-3.65,7)};
\addplot[black!62, thick, forget plot] coordinates {(-4.12,8) (0.96,8)};
\addplot[only marks, mark=*, mark size=1.9pt, black!62, forget plot] coordinates {(-1.48,8)};
\addplot[black!62, thick, forget plot] coordinates {(-7.69,9) (2.65,9)};
\addplot[only marks, mark=*, mark size=1.9pt, black!62, forget plot] coordinates {(-2.42,9)};
\addplot[c3, thick, forget plot] coordinates {(-19.99,10) (-8.45,10)};
\addplot[only marks, mark=*, mark size=1.9pt, c3, forget plot] coordinates {(-14.07,10)};
\addplot[black!62, thick, forget plot] coordinates {(-10.26,11) (1.79,11)};
\addplot[only marks, mark=*, mark size=1.9pt, black!62, forget plot] coordinates {(-4.22,11)};
\addplot[black!62, thick, forget plot] coordinates {(-7.05,12) (4.10,12)};
\addplot[only marks, mark=*, mark size=1.9pt, black!62, forget plot] coordinates {(-1.51,12)};
\addplot[black!62, thick, forget plot] coordinates {(-10.53,13) (0.13,13)};
\addplot[only marks, mark=*, mark size=1.9pt, black!62, forget plot] coordinates {(-5.06,13)};
\addplot[black!62, thick, forget plot] coordinates {(-0.20,14) (10.06,14)};
\addplot[only marks, mark=*, mark size=1.9pt, black!62, forget plot] coordinates {(4.91,14)};
\addplot[black!62, thick, forget plot] coordinates {(0.39,15) (11.14,15)};
\addplot[only marks, mark=*, mark size=1.9pt, black!62, forget plot] coordinates {(5.68,15)};
\addplot[black!62, thick, forget plot] coordinates {(-6.48,16) (1.71,16)};
\addplot[only marks, mark=*, mark size=1.9pt, black!62, forget plot] coordinates {(-2.33,16)};
\addplot[black!62, thick, forget plot] coordinates {(-9.87,17) (-0.17,17)};
\addplot[only marks, mark=*, mark size=1.9pt, black!62, forget plot] coordinates {(-4.94,17)};
\addplot[c3, thick, forget plot] coordinates {(-9.57,18) (-1.32,18)};
\addplot[only marks, mark=*, mark size=1.9pt, c3, forget plot] coordinates {(-5.29,18)};
\addplot[black!62, thick, forget plot] coordinates {(-4.96,19) (1.97,19)};
\addplot[only marks, mark=*, mark size=1.9pt, black!62, forget plot] coordinates {(-1.37,19)};
\addplot[c3, thick, forget plot] coordinates {(-8.51,20) (-1.29,20)};
\addplot[only marks, mark=*, mark size=1.9pt, c3, forget plot] coordinates {(-4.73,20)};
\addplot[black!62, thick, forget plot] coordinates {(-5.49,21) (1.96,21)};
\addplot[only marks, mark=*, mark size=1.9pt, black!62, forget plot] coordinates {(-1.74,21)};
\addplot[black!62, thick, forget plot] coordinates {(-4.43,22) (2.11,22)};
\addplot[only marks, mark=*, mark size=1.9pt, black!62, forget plot] coordinates {(-1.06,22)};
\addplot[black!62, thick, forget plot] coordinates {(-11.48,23) (0.18,23)};
\addplot[only marks, mark=*, mark size=1.9pt, black!62, forget plot] coordinates {(-5.55,23)};
\addplot[c3, thick, forget plot] coordinates {(-14.18,24) (-2.08,24)};
\addplot[only marks, mark=*, mark size=1.9pt, c3, forget plot] coordinates {(-8.12,24)};
\addplot[c3, thick, forget plot] coordinates {(-15.26,25) (-2.68,25)};
\addplot[only marks, mark=*, mark size=1.9pt, c3, forget plot] coordinates {(-9.01,25)};
\addplot[black!62, thick, forget plot] coordinates {(-3.24,26) (6.77,26)};
\addplot[only marks, mark=*, mark size=1.9pt, black!62, forget plot] coordinates {(1.73,26)};
\addplot[black!62, thick, forget plot] coordinates {(-11.80,27) (-0.84,27)};
\addplot[only marks, mark=*, mark size=1.9pt, black!62, forget plot] coordinates {(-6.29,27)};
\addplot[black!62, thick, forget plot] coordinates {(-5.75,28) (5.24,28)};
\addplot[only marks, mark=*, mark size=1.9pt, black!62, forget plot] coordinates {(-0.23,28)};
\addplot[black!62, thick, forget plot] coordinates {(-2.28,29) (8.67,29)};
\addplot[only marks, mark=*, mark size=1.9pt, black!62, forget plot] coordinates {(3.17,29)};
\addplot[black!62, thick, forget plot] coordinates {(-9.96,30) (-0.66,30)};
\addplot[only marks, mark=*, mark size=1.9pt, black!62, forget plot] coordinates {(-5.19,30)};
\addplot[c3, thick, forget plot] coordinates {(-13.31,31) (-2.23,31)};
\addplot[only marks, mark=*, mark size=1.9pt, c3, forget plot] coordinates {(-7.74,31)};
\addplot[black!62, thick, forget plot] coordinates {(-11.07,32) (1.10,32)};
\addplot[only marks, mark=*, mark size=1.9pt, black!62, forget plot] coordinates {(-4.95,32)};
\addplot[black!62, thick, forget plot] coordinates {(-1.20,33) (8.87,33)};
\addplot[only marks, mark=*, mark size=1.9pt, black!62, forget plot] coordinates {(3.91,33)};
\addplot[black!62, thick, forget plot] coordinates {(-11.47,34) (-1.15,34)};
\addplot[only marks, mark=*, mark size=1.9pt, black!62, forget plot] coordinates {(-6.25,34)};
\addplot[black!62, thick, forget plot] coordinates {(-4.95,35) (4.88,35)};
\addplot[only marks, mark=*, mark size=1.9pt, black!62, forget plot] coordinates {(-0.09,35)};
\addplot[black!62, thick, forget plot] coordinates {(-3.02,36) (6.83,36)};
\addplot[only marks, mark=*, mark size=1.9pt, black!62, forget plot] coordinates {(1.94,36)};
\end{axis}
\end{tikzpicture}
\caption{Difference in accuracy between each method and self-consistency at equal generated-token cost, for every method and setting. Points left of the dashed line mean the method did worse than simply drawing more samples for the same tokens. Bars are $95\%$ paired bootstrap intervals over questions. \textbf{Coloured} intervals are the comparisons that remain significant after Holm correction within their setting; \textbf{grey} intervals are not significant. All of the significant ones fall on the left of the line.}
\label{fig:forest}\end{figure}

Table~\ref{tab:main} and Figure~\ref{fig:forest} give the answer for all 36 method--setting comparisons. After Holm correction within each setting, 0 are significantly \emph{better} than self-consistency at matched token cost, 10 are significantly \emph{worse}, and 26 are statistically indistinguishable from it.

Counting significant cells is a blunt summary, and on its own it understates what the table shows. 30 of the 36 point estimates are negative and 15 comparisons have a raw $95\%$ interval excluding zero before correction. But the interesting structure is not in the totals.

\paragraph{The methods that ask the model to judge its own work lose; the ones that count do not.} Grouping by what a method does with its extra tokens separates the table cleanly. The three methods in which the model assesses or rewrites its own output (Self-Refine, Reflexion (forced), Best-of-$N$) are below the cost-matched baseline in \emph{all 18} of their comparisons. The methods that add no self-assessment are at chance (14 of 20 negative, $p=0.12$). We must be candid that this grouping was formed \emph{after} we corrected the Best-of-$N$ scoring error described in Appendix~\ref{app:repro}, so we report it as a description of the pattern rather than as a hypothesis test, and we give the mechanism behind it in Section~\ref{sec:results} below rather than resting on a $p$-value.

\paragraph{The same holds for the grouping fixed in advance.} Pooling all 36 differences, 30 of 36 are negative, which also points the same way ($p<0.001$). Splitting instead by what the methods do sharpens it. The three methods that spend their budget on repeated passes over their own work (Self-Refine, Reflexion (forced), Multi-Agent Debate) are below the equal-cost baseline in every one of their 16 comparisons ($p=0.0000$). The remaining methods are at chance: 14 of 20 negative ($p=0.12$).

That sign test treats the 16 comparisons as independent, which they are not: within a setting they share a baseline pool and a question sample. A cluster-robust version asks instead whether \emph{all} the iterative methods come out negative within a setting, and counts settings. That happens in 6 of 6 settings, giving an exact $p$ of 0.0156. That is weaker, as it should be with only four independent units, but it points the same way. We treat the clustered figure as the honest one.

The methods that are significantly worse are: Best-of-$N$ on Qwen2.5-1.5B/GSM8K (-7.7~pp, 95\% CI [-13.3, -2.2]); Best-of-$N$ on Qwen2.5-1.5B/MATH-500 (-8.1~pp, 95\% CI [-14.2, -2.1]); Reflexion (forced) on Qwen2.5-1.5B/MATH-500 (-9.0~pp, 95\% CI [-15.3, -2.7]); Reflexion (forced) on Qwen2.5-3B/GSM8K (-5.3~pp, 95\% CI [-9.6, -1.3]); Self-Refine on Qwen2.5-3B/GSM8K (-4.7~pp, 95\% CI [-8.5, -1.3]); Best-of-$N$ on Qwen2.5-3B/MATH-500 (-14.1~pp, 95\% CI [-20.0, -8.4]); Reflexion (forced) on Qwen2.5-7B/GSM8K (-5.5~pp, 95\% CI [-10.2, -1.2]); Self-Refine on Qwen2.5-7B/GSM8K (-3.6~pp, 95\% CI [-6.9, -0.9]); Reflexion (forced) on Qwen2.5-7B/MATH-500 (-10.1~pp, 95\% CI [-15.0, -5.7]); Self-Refine on Qwen2.5-7B/MATH-500 (-6.3~pp, 95\% CI [-11.1, -2.0]).

\begin{table}[htbp]\centering\small
\setlength{\tabcolsep}{4pt}
\begin{tabular}{@{}llrrrr@{\;}l r@{}}
\toprule
Setting & Method & Acc.\ & Tokens & SC acc.\ & \multicolumn{2}{c}{$\Delta$ (pp), 95\% CI} & $p_{\text{Holm}}$\\
\midrule
\emph{1.5B/GSM8K} & Chain-of-Thought & 72.0 & 292 & 70.1 & +1.9 & \footnotesize[-3.0,+6.8] & 0.869\\
 & Plan-and-Solve & 70.0 & 319 & 70.1 & -0.1 & \footnotesize[-5.0,+4.9] & 0.989\\
 & Self-Refine & 72.0 & 1784 & 78.3 & -6.3 & \footnotesize[-11.5,-1.2] & 0.097\\
 & Reflexion & 74.0 & 323 & 70.1 & +3.9 & \footnotesize[-1.2,+8.9] & 0.441\\
 & Reflexion (forced) & 71.3 & 1197 & 76.3 & -5.0 & \footnotesize[-11.1,+1.1] & 0.441\\
 & Best-of-$N$ (self-verify) & 71.3 & 2369 & 79.1 & -7.7 & \footnotesize[-13.3,-2.2] & 0.036\\
 & Multi-Agent Debate & 74.0 & 2539 & 79.2 & -5.2 & \footnotesize[-10.0,-0.7] & 0.110\\
\midrule
\emph{1.5B/MATH-500} & Chain-of-Thought & 47.3 & 566 & 44.2 & +3.2 & \footnotesize[-2.3,+8.7] & 0.775\\
 & Plan-and-Solve & 44.0 & 591 & 44.2 & -0.2 & \footnotesize[-5.7,+5.2] & 1.000\\
 & Self-Refine & 47.3 & 3467 & 53.6 & -6.3 & \footnotesize[-11.8,-0.8] & 0.105\\
 & Reflexion & 46.0 & 608 & 44.3 & +1.7 & \footnotesize[-3.2,+6.8] & 1.000\\
 & Reflexion (forced) & 43.3 & 2827 & 52.3 & -9.0 & \footnotesize[-15.3,-2.7] & 0.041\\
 & Best-of-$N$ (self-verify) & 46.7 & 4320 & 54.8 & -8.1 & \footnotesize[-14.2,-2.1] & 0.041\\
 & Multi-Agent Debate & 49.3 & 4493 & 54.9 & -5.6 & \footnotesize[-11.5,+0.2] & 0.229\\
\midrule
\emph{3B/GSM8K} & Chain-of-Thought & 83.3 & 295 & 84.4 & -1.1 & \footnotesize[-4.4,+2.1] & 1.000\\
 & Plan-and-Solve & 82.7 & 328 & 84.4 & -1.7 & \footnotesize[-5.5,+2.0] & 1.000\\
 & Self-Refine & 83.3 & 1436 & 88.1 & -4.7 & \footnotesize[-8.5,-1.3] & 0.029\\
 & Reflexion & 84.0 & 712 & 85.4 & -1.4 & \footnotesize[-5.0,+2.0] & 1.000\\
 & Reflexion (forced) & 83.3 & 1956 & 88.6 & -5.3 & \footnotesize[-9.6,-1.3] & 0.035\\
 & Best-of-$N$ (self-verify) & 84.0 & 2404 & 88.9 & -4.9 & \footnotesize[-9.9,-0.2] & 0.198\\
 & Multi-Agent Debate & 86.7 & 2681 & 89.0 & -2.3 & \footnotesize[-6.5,+1.7] & 1.000\\
\midrule
\emph{3B/MATH-500} & Chain-of-Thought & 61.3 & 550 & 55.7 & +5.7 & \footnotesize[+0.4,+11.1] & 0.204\\
 & Plan-and-Solve & 60.7 & 588 & 55.8 & +4.9 & \footnotesize[-0.2,+10.1] & 0.279\\
 & Self-Refine & 58.7 & 2705 & 63.7 & -5.1 & \footnotesize[-10.5,+0.1] & 0.279\\
 & Reflexion & 62.0 & 2592 & 63.5 & -1.5 & \footnotesize[-7.0,+4.1] & 0.712\\
 & Reflexion (forced) & 60.7 & 3289 & 64.9 & -4.2 & \footnotesize[-10.3,+1.8] & 0.518\\
 & Best-of-$N$ (self-verify) & 52.0 & 4391 & 66.1 & -14.1 & \footnotesize[-20.0,-8.4] & $<$0.001\\
 & Multi-Agent Debate & 64.0 & 4814 & 66.4 & -2.4 & \footnotesize[-7.7,+2.7] & 0.712\\
\midrule
\emph{7B/GSM8K} & Chain-of-Thought & 88.7 & 291 & 90.1 & -1.5 & \footnotesize[-4.1,+1.0] & 0.254\\
 & Self-Refine & 88.7 & 1374 & 92.3 & -3.6 & \footnotesize[-6.9,-0.9] & 0.026\\
 & Reflexion (forced) & 87.3 & 1885 & 92.9 & -5.5 & \footnotesize[-10.2,-1.2] & 0.032\\
 & Best-of-$N$ (self-verify) & 90.7 & 2282 & 93.2 & -2.6 & \footnotesize[-5.8,+0.4] & 0.190\\
\midrule
\emph{7B/MATH-500} & Chain-of-Thought & 66.7 & 543 & 67.9 & -1.2 & \footnotesize[-5.3,+2.7] & 0.558\\
 & Self-Refine & 66.7 & 2627 & 73.0 & -6.3 & \footnotesize[-11.1,-2.0] & 0.018\\
 & Reflexion (forced) & 63.3 & 3206 & 73.5 & -10.1 & \footnotesize[-15.0,-5.7] & $<$0.001\\
 & Best-of-$N$ (self-verify) & 71.3 & 4335 & 73.7 & -2.4 & \footnotesize[-6.2,+1.2] & 0.411\\
\bottomrule
\end{tabular}

\caption{Accuracy, mean generated tokens per question, the accuracy of self-consistency at the same token cost, and their paired difference with a $95\%$ bootstrap interval. $p$ values are Holm-corrected across the methods within each setting.}
\label{tab:main}\end{table}

\subsection{The cost--accuracy picture}

\begin{figure}[htbp]\centering
\input{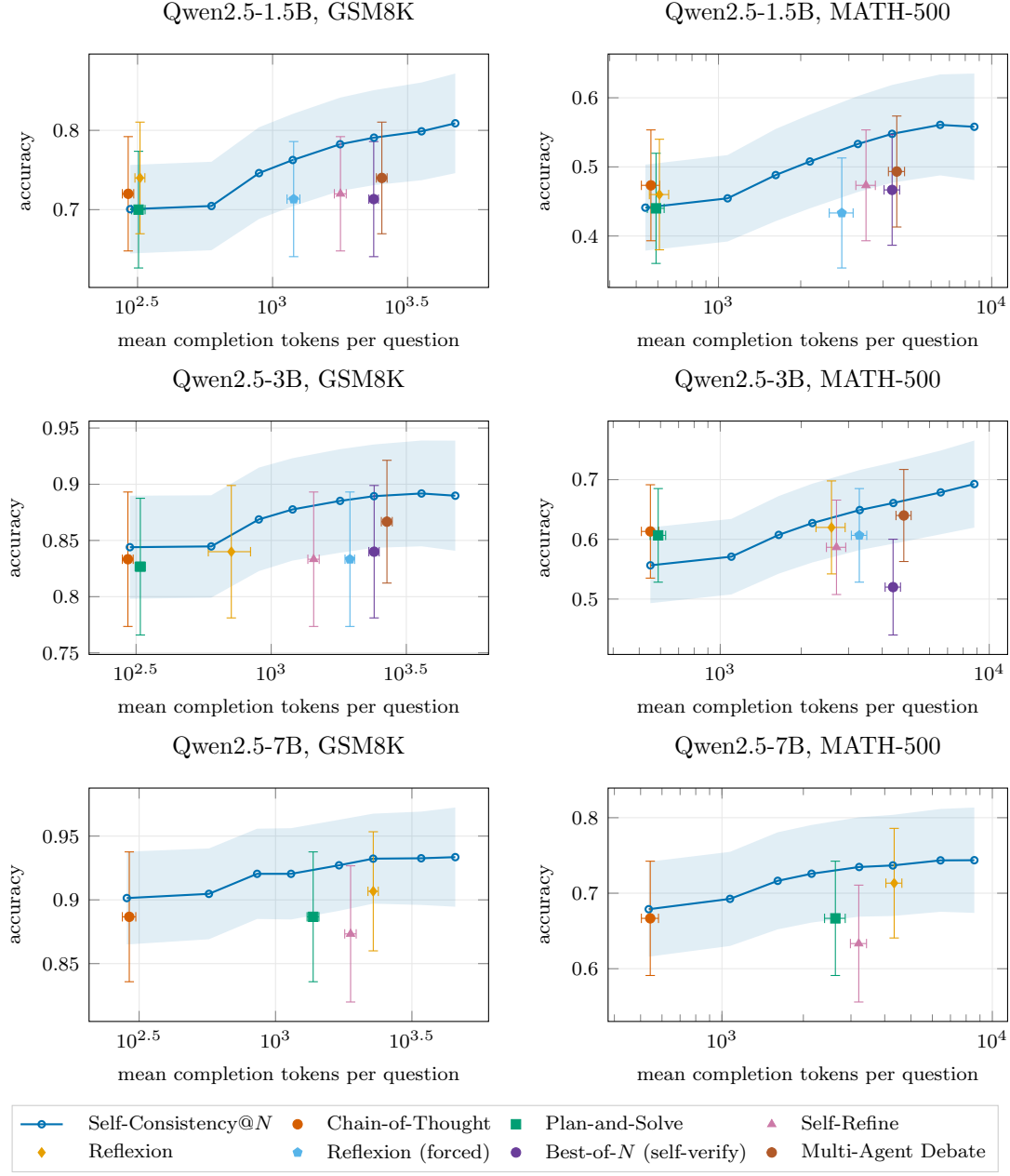}
\\[4pt]
\begin{tikzpicture}
\begin{axis}[hide axis, xmin=0, xmax=1, ymin=0, ymax=1,
  width=0.9\textwidth, height=1.1cm, scale only axis,
  legend columns=4, legend style={font=\scriptsize, draw=gray!40,
  column sep=8pt, /tikz/every even column/.append style={column sep=6pt}},
  legend cell align=left]
\addlegendimage{thick, color=scblue, mark=o, mark size=1.3pt}
\addlegendentry{Self-Consistency@$N$}
\addlegendimage{only marks, mark=*, mark size=1.9pt, color=c1}
\addlegendentry{Chain-of-Thought}
\addlegendimage{only marks, mark=square*, mark size=1.9pt, color=c2}
\addlegendentry{Plan-and-Solve}
\addlegendimage{only marks, mark=triangle*, mark size=1.9pt, color=c3}
\addlegendentry{Self-Refine}
\addlegendimage{only marks, mark=diamond*, mark size=1.9pt, color=c4}
\addlegendentry{Reflexion}
\addlegendimage{only marks, mark=pentagon*, mark size=1.9pt, color=c5}
\addlegendentry{Reflexion (forced)}
\addlegendimage{only marks, mark=otimes*, mark size=1.9pt, color=c6}
\addlegendentry{Best-of-$N$ (self-verify)}
\addlegendimage{only marks, mark=oplus*, mark size=1.9pt, color=c7}
\addlegendentry{Multi-Agent Debate}
\end{axis}
\end{tikzpicture}
\caption{Accuracy against mean generated tokens per question. The shaded band is a $95\%$ interval on the self-consistency curve; error bars on the markers are $95\%$ intervals on both axes.}
\label{fig:frontier}\end{figure}

Figure~\ref{fig:frontier} plots accuracy against mean generated tokens. The line is self-consistency as $N$ grows; each marker is one method at its own measured cost. A method is worth its budget only if it sits above the line.

On Qwen2.5-1.5B/GSM8K, self-consistency rises from 70.1\% at $N=1$ (297 tokens) to 80.9\% at $N=16$ (4749 tokens).
On Qwen2.5-1.5B/MATH-500, self-consistency rises from 44.1\% at $N=1$ (540 tokens) to 55.8\% at $N=16$ (8636 tokens).
On Qwen2.5-3B/GSM8K, self-consistency rises from 84.4\% at $N=1$ (300 tokens) to 89.0\% at $N=16$ (4799 tokens).
On Qwen2.5-3B/MATH-500, self-consistency rises from 55.7\% at $N=1$ (550 tokens) to 69.3\% at $N=16$ (8803 tokens).
On Qwen2.5-7B/GSM8K, self-consistency rises from 90.1\% at $N=1$ (285 tokens) to 93.4\% at $N=16$ (4560 tokens).
On Qwen2.5-7B/MATH-500, self-consistency rises from 67.9\% at $N=1$ (536 tokens) to 74.4\% at $N=16$ (8578 tokens).

\subsection{Where the loss comes from: judging versus counting}

The comparisons above match a method against a \emph{different} procedure at equal cost, so a sceptic can always ask whether the gap is really about the mechanism. Best-of-$N$ lets us remove that doubt entirely. It draws $N=8$ samples and then asks the model which one is best. We can take those same eight samples and instead simply count which answer appears most often. The samples, the tokens and the model are identical. The only difference is whether the winner is picked by the model or by a tally.

\begin{center}\small\begin{tabular}{lrrr@{\;}lr}
\toprule
Setting & judges & counting & \multicolumn{2}{c}{difference, 95\% CI} & $p$\\
\midrule
Qwen2.5-1.5B/GSM8K & 71.3\% & 79.3\% & -8.0 pp & \footnotesize[-13.3, -3.3] & 0.001\\
Qwen2.5-1.5B/MATH-500 & 46.7\% & 58.0\% & -11.3 pp & \footnotesize[-17.3, -6.0] & $<$0.001\\
Qwen2.5-3B/GSM8K & 84.0\% & 89.3\% & -5.3 pp & \footnotesize[-10.7, -0.7] & 0.038\\
Qwen2.5-3B/MATH-500 & 52.0\% & 69.3\% & -17.3 pp & \footnotesize[-24.0, -11.3] & $<$0.001\\
Qwen2.5-7B/GSM8K & 90.7\% & 92.7\% & -2.0 pp & \footnotesize[-5.3, +0.7] & 0.241\\
Qwen2.5-7B/MATH-500 & 71.3\% & 72.7\% & -1.3 pp & \footnotesize[-4.7, +2.0] & 0.538\\
\bottomrule\end{tabular}\end{center}

Counting wins in every setting, by between 1.3 and 17.3 percentage points. The verifier agrees with the majority on 63--95\% of questions; it is on the remainder that it loses ground. This is the cleanest statement of the paper's finding: holding the samples themselves fixed, asking the model to evaluate its own candidates is worse than not asking it.

\paragraph{Does it close with scale?} This is the obvious objection, that self-evaluation simply needs a bigger model, so we ran the same comparison across 1.5B and 3B and 7B. The penalty shrinks steadily, and so does the disagreement behind it: the model's choice matches the majority answer more and more often as the model grows.

Statistically the picture is clean. Below $7$B the gap is significant in 4 of 4 settings. At $7$B it is significant in 0 of 2, and both intervals contain zero (Qwen2.5-7B/GSM8K -2.0 pp, CI [-5.3, +0.7]; Qwen2.5-7B/MATH-500 -1.3 pp, CI [-4.7, +2.0]). We therefore do not claim that counting still beats judging at $7$B; we claim that it does so below $7$B, and that by $7$B the two are not distinguishable with 150 questions. The intervals are narrow enough that any remaining penalty at $7$B is small rather than merely unmeasured.

That crossover is itself informative. \citet{zhang2025selfverify} report that a verifier \emph{trained} to judge does beat majority voting at about this scale. Our numbers say an untrained verifier only reaches parity there. Read together, the training rather than the judging appears to be what makes a verifier worth its tokens.
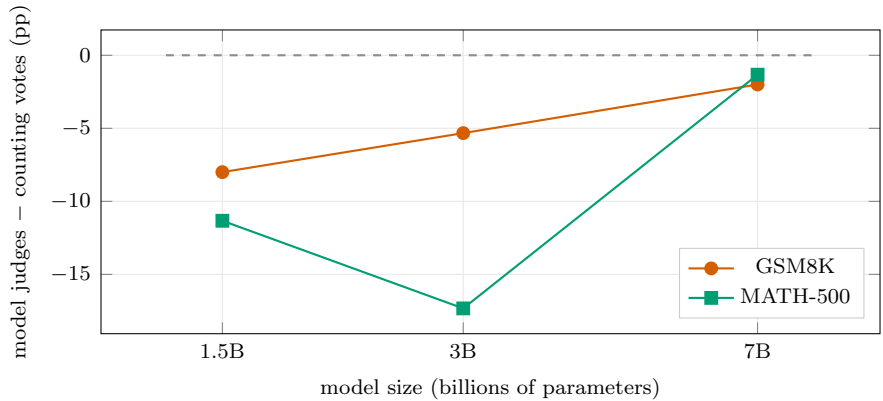
\begin{figure}[htbp]\centering
\begin{tikzpicture}
\begin{axis}[width=0.72\textwidth, height=5.6cm,
  xlabel={model size (billions of parameters)},
  ylabel={model judges $-$ counting votes (pp)},
  xmode=log, log basis x=2, xtick={1.5,3,7},
  xticklabels={1.5B,3B,7B},
  grid=major, grid style={gray!18},
  tick label style={font=\scriptsize}, label style={font=\scriptsize},
  legend style={font=\scriptsize, at={(0.98,0.05)}, anchor=south east,
                draw=gray!40}]
\addplot[black!45, dashed, thick, forget plot, samples=2, domain=1.275:8.260] {0};
\addplot[thick, color=c1, mark=*, mark size=2.2pt] coordinates {(1.5,-8.00) (3.0,-5.33) (7.0,-2.00)};
\addlegendentry{GSM8K}
\addplot[thick, color=c2, mark=square*, mark size=2.2pt] coordinates {(1.5,-11.33) (3.0,-17.33) (7.0,-1.33)};
\addlegendentry{MATH-500}
\end{axis}
\end{tikzpicture}
\caption{The cost of asking the model to judge, against model size. Both quantities are measured on the \emph{same} sampled solutions, so the vertical axis isolates the value of the model's selection step from everything else. Points below the dashed line mean counting votes would have been better.}
\label{fig:scale}\end{figure}

Debate shows the same tension from the other side. After two rounds its three agents agree unanimously on 48--91\% of questions, so most of the time the debate has converged and the extra rounds bought agreement rather than accuracy. Debate is also the method that loses \emph{least} to the baseline, which fits: it is the one whose final step is a vote rather than a judgement.

\subsection{Is the comparison just a temperature effect?}

Greedy chain-of-thought runs at temperature $0$ while the baseline samples at $0.7$, so we check the two directly. Self-consistency at $N=1$ is a single sample at $0.7$ and costs about what one greedy chain costs, which isolates temperature from everything else.

On Qwen2.5-1.5B/GSM8K the difference (greedy minus sampled) is +1.9~pp, $95\%$ CI $[-3.0, +6.8]$, $p=0.428$.
On Qwen2.5-1.5B/MATH-500 the difference (greedy minus sampled) is +3.2~pp, $95\%$ CI $[-2.2, +8.8]$, $p=0.248$.
On Qwen2.5-3B/GSM8K the difference (greedy minus sampled) is -1.1~pp, $95\%$ CI $[-4.5, +2.1]$, $p=0.537$.
On Qwen2.5-3B/MATH-500 the difference (greedy minus sampled) is +5.7~pp, $95\%$ CI $[+0.5, +11.0]$, $p=0.031$.
On Qwen2.5-7B/GSM8K the difference (greedy minus sampled) is -1.5~pp, $95\%$ CI $[-4.1, +1.0]$, $p=0.269$.
On Qwen2.5-7B/MATH-500 the difference (greedy minus sampled) is -1.2~pp, $95\%$ CI $[-5.2, +2.7]$, $p=0.538$.

\subsection{Robustness}

\paragraph{Unparseable answers.} On the GSM8K settings no response failed to parse. On MATH-500 some did: the highest rate is 16.0\% (Reflexion on Qwen2.5-1.5B/MATH-500). We do not wave this away. An answer we cannot parse is scored wrong, so it counts against the method that produced it.

The cause is the $1{,}024$-token cap on a single generation: long MATH-500 solutions are cut off before the model reaches its final boxed answer. The cap applies equally to every method, but its \emph{consequences} do not. A method that draws several samples and votes needs only some of them to finish, so Best-of-$N$ and debate lose well under one percent of answers, whereas single-shot methods lose ten percent or more. That robustness is a genuine property of sampling rather than an artefact, but it is entangled with our choice of cap, so we check it directly below.

\paragraph{Restricting to questions every method answered (Qwen2.5-1.5B/MATH-500).} 84 of the 150 questions produced a parseable answer from every method and every baseline sample. Repeating the whole comparison on that subset moves each estimated difference by at most 5.2 percentage points. The verdict is \emph{not} stable for every method: Best-of-$N$ moves from -8.1~pp ($p=0.041$) to -6.9~pp ($p=0.564$), ceasing to be significant. We flag this rather than choose whichever analysis we prefer. The subset is not a random sample. It excludes exactly the questions on which some method ran out of tokens, which are the longer and harder ones. Neither analysis is therefore automatically the correct one, and a difference that appears only under one of them should be treated as suggestive rather than established.

\paragraph{Restricting to questions every method answered (Qwen2.5-3B/MATH-500).} 101 of the 150 questions produced a parseable answer from every method and every baseline sample. Repeating the whole comparison on that subset moves each estimated difference by at most 5.2 percentage points. The verdict is \emph{not} stable for every method: Plan-and-Solve moves from +4.9~pp ($p=0.279$) to +8.5~pp ($p=0.018$), becoming significant. We flag this rather than choose whichever analysis we prefer. The subset is not a random sample. It excludes exactly the questions on which some method ran out of tokens, which are the longer and harder ones. Neither analysis is therefore automatically the correct one, and a difference that appears only under one of them should be treated as suggestive rather than established.

\paragraph{Restricting to questions every method answered (Qwen2.5-7B/MATH-500).} 109 of the 150 questions produced a parseable answer from every method and every baseline sample. Repeating the whole comparison on that subset moves each estimated difference by at most 5.0 percentage points. The verdict is \emph{not} stable for every method: Best-of-$N$ moves from -2.4~pp ($p=0.411$) to +2.7~pp ($p=0.024$), becoming significant; Self-Refine moves from -6.3~pp ($p=0.018$) to -4.5~pp ($p=0.058$), ceasing to be significant. We flag this rather than choose whichever analysis we prefer. The subset is not a random sample. It excludes exactly the questions on which some method ran out of tokens, which are the longer and harder ones. Neither analysis is therefore automatically the correct one, and a difference that appears only under one of them should be treated as suggestive rather than established.

\paragraph{How often Reflexion actually reflects.} Reflexion retries only when the model judges its own answer wrong, so the rate at which that judgement fires determines whether the method is doing anything at all. It varies enormously with model size:

\begin{center}\small\begin{tabular}{lrr}
\toprule
Setting & stopped immediately & mean rounds used (max 3)\\
\midrule
Qwen2.5-1.5B/GSM8K & 100\% & 0.00\\
Qwen2.5-1.5B/MATH-500 & 100\% & 0.00\\
Qwen2.5-3B/GSM8K & 77\% & 0.60\\
Qwen2.5-3B/MATH-500 & 31\% & 1.82\\
\bottomrule\end{tabular}\end{center}

On Qwen2.5-1.5B/GSM8K and Qwen2.5-1.5B/MATH-500 the self-assessment never fired once: the model called its first answer correct on every question, and what was measured under the name Reflexion was a single chain of thought. Its apparently favourable score is a consequence of that, not evidence that reflection helps. The forced variant, which performs the same three rounds regardless, is the comparison that actually tests the mechanism.

\paragraph{Correcting across all settings.} Applying Holm across all 36 comparisons at once rather than within each setting leaves 2 significant at the $5\%$ level, compared with 10 under the within-setting correction.

\paragraph{Does difficulty change the verdict?} MATH-500 labels each problem with a level from 1 to 5. Splitting Qwen2.5-1.5B/MATH-500 by level gives the following mean differences against the cost-matched baseline. These bins are small, so we report them as exploratory and do not correct them for repeated testing or draw conclusions from individual cells.

\begin{center}\small\begin{tabular}{lrrr}
\toprule
Method & easy (L1-2) ($n=38$) & medium (L3) ($n=30$) & hard (L4-5) ($n=82$)\\
\midrule
Chain-of-Thought & +2.5 & +6.5 & +2.1\\
Plan-and-Solve & +5.1 & +6.5 & -5.2\\
Self-Refine & -3.2 & -9.4 & -6.5\\
Reflexion & +2.5 & +6.4 & -0.4\\
Reflexion (forced) & -4.1 & -15.0 & -9.2\\
Best-of-$N$ & +1.2 & -21.7 & -7.0\\
Multi-Agent Debate & +0.9 & -8.7 & -7.1\\
\bottomrule\end{tabular}\end{center}

\paragraph{Monte-Carlo stability.} Re-running the whole analysis with a different random seed for the subsampling changes each estimated difference by at most 0.36 percentage points (mean 0.12), which is far smaller than the confidence intervals and does not alter any conclusion.

\section{What this means, and what it does not}
\label{sec:discussion}

\subsection{Reading the result correctly}

The claim we can support is narrow and worth stating precisely, and it is
weaker than either ``these methods work'' or ``these methods do not work''.

What we observe is that no method is significantly better than plain repeated
sampling once the two are given the same number of generated tokens, and that
six are significantly worse. All six ask the model to judge or rewrite its own
output.

Two different ways of grouping the methods tell the same story, which is worth
separating carefully because one of them was chosen after we had seen results.
The grouping we fixed in advance was ``methods that make repeated passes over
their own work'', which covers Self-Refine, forced Reflexion and debate. The
grouping we formed afterwards, once the Best-of-$N$ scoring error was
corrected, was ``methods in which the model assesses its own output'', which
replaces debate with Best-of-$N$. Both come out below the equal-cost baseline
in all twelve of their comparisons. We would not lean on either grouping by
itself, and we do not need to, because the Best-of-$N$ comparison in
Section~\ref{sec:results} makes the same point without grouping anything.

Counting settings rather than comparisons, so that results sharing a question
sample are not counted as independent evidence, the pattern is suggestive
rather than conclusive ($p = 0.06$ on four settings). Our intervals are wide
enough that we cannot rule out modest real benefits for any individual method,
and we do not claim the true effects are zero.

The right reading is therefore about the \emph{comparison}, not about the
methods. Reporting a gain over one chain of thought does not establish that a
method's mechanism is responsible for that gain, because a cheaper and more
boring use of the same tokens is available and is rarely reported alongside
it. This distinction matters in practice: with a fixed budget, the question
is not ``does self-criticism help?'' but ``does self-criticism help more than
drawing more samples for the same tokens?'' Those are different questions,
and only the second one has an actionable answer.

\subsection{A method can pass by not running}

The clearest lesson from our runs is not about any method's score. Reflexion,
implemented exactly as specified, decides for itself when to stop: it reflects
and retries only when the model judges its own answer wrong. On Qwen2.5-1.5B
the model judged itself right on \emph{every} question in both benchmarks, so
the loop exited immediately every time, and what we were measuring was a
single chain of thought wearing Reflexion's name. It looked like one of the
better methods precisely because it was one of the cheapest. It never did
the thing it exists to do. On Qwen2.5-3B the same code behaves quite
differently, retrying on roughly a quarter of GSM8K questions and on most
MATH-500 ones, which is why its cost there is several times higher. The same
method, unchanged, is a different experiment on a different model.

Had we not instrumented how often the self-assessment fires, we would have
reported that number as a result about reflection. When we force the loop to
run, the score moves in the opposite direction and lands below the baseline.
The general point is that a method with a self-triggered control flow can
silently degenerate into a cheaper method, and a benchmark that reports only
final accuracy cannot tell the difference. Any evaluation of adaptive methods
should report how often the adaptive part actually engages; we would treat a
paper that does not as having left a hole in its evidence, and we report the
rate for our own runs in Section~\ref{sec:results}.

The underlying gap, between what a procedure guarantees on paper and what the
running artefact does, is not particular to language models. It appears
wherever a stated property is checked against a specification rather than
against measurements of a deployment. In consensus protocols, for instance, an
ordering rule that is fair by specification has been shown to grant some
participants a systematic advantage once the implementation is measured
\citep{mirzaei2026fair}. The parallel is only structural, since nothing else
about that setting resembles ours, but it is the same failure of inference:
reading a property off the design instead of off the artefact.

\subsection{Why the gap closes, and what that implies above $7$B}

The obvious question about a shrinking gap is whether it eventually reverses.
If a model large enough could judge better than a tally, then everything here
is a statement about small models and nothing more. The data contain a
diagnostic that speaks to this directly, because the verifier can only differ
from the majority on the questions where the two disagree, and on those
questions we can ask which one is right.

Two things happen as the model grows, and only one of them is what a reversal
would need. Disagreements become much rarer: the verifier departs from the
majority on $30$, then $17$, then $8$ of $150$ GSM8K questions at $1.5$B, $3$B
and $7$B, and on $55$, $48$ and $24$ of MATH-500. But when it does depart, it
is still usually wrong. Among the disagreements where exactly one of the two is
correct, the verifier is the correct one in $12.5\%$, $21.4\%$ and $20.0\%$ of
cases on GSM8K, and $9.5\%$, $3.6\%$ and $33.3\%$ on MATH-500. Every one of
those is below the $50\%$ that would make judging and counting equally good on
the cases where they differ.

So the gap closes because the verifier increasingly agrees with the tally, not
because it becomes a better judge than the tally. That is convergence from
below rather than an approach to a crossing. For judging to overtake counting,
the override accuracy would have to climb past one half, and we see no trend
towards it over the range we can measure. We say this as a falsifiable
prediction rather than a result: the counts at $7$B are small, eight and
twenty-four disagreements, so the intervals are wide, and a larger model could
in principle behave differently. But the mechanism that would have to change is
now named and cheap to check, which is more useful than an assertion that the
finding does or does not extrapolate.

This also fits the one result in the literature that appears to point the other
way. \citet{zhang2025selfverify} find that a verifier \emph{trained} to judge
does beat majority voting at about this scale. Training is exactly the
intervention that would raise override accuracy, which is the quantity our
untrained verifier fails to improve. The two findings are consistent, and
together they locate the useful ingredient in the training rather than in the
act of verifying.

\subsection{Choosing a cost measure}

We count generated tokens. Three other measures are defensible, and each
would shift the picture:

\begin{itemize}
\itemsep2pt
\item \textbf{Input tokens.} Some methods re-read long contexts. In debate,
      each agent reads every other agent's full solution on every round, so it
      necessarily consumes more input tokens than the baseline, which sends
      only the question. We did not record input tokens and so cannot put a
      number on this, but the direction is fixed by how the methods are built:
      under a cost measure that charges for input, debate looks \emph{worse}
      than we report, not better.
\item \textbf{Wall-clock latency.} The sampling baseline is embarrassingly
      parallel: $16$ chains can be generated simultaneously. Self-Refine and
      Reflexion are strictly sequential; each round must finish before the
      next begins. Under a latency measure with enough hardware to run
      samples in parallel, the sequential methods look worse still.
\item \textbf{Money.} Priced per token by a commercial API, the ranking
      matches our token-based ranking closely, since output tokens dominate.
\end{itemize}

We chose generated tokens because it is the measure least dependent on
deployment details, and because it is the one on which the sequential methods
look \emph{best}. Our comparison is therefore conservative with respect to
the methods we are testing.

\subsection{Limitations}

\paragraph{Model scale.} The cost-matched comparison covers $1.5$B and $3$B
parameter models. The Best-of-$N$ comparison, which is the cheapest to run,
also covers $7$B, and it is there that the effect disappears. Larger models
are better at judging their own work, and that is exactly what our scaling
numbers show. We cannot go beyond $7$B on the hardware available, so whether
the two curves cross again at frontier scale, with judging eventually beating
counting, is untested. Our data locate the point where they draw level, not
what happens above it.

\paragraph{Quantisation.} Weights are stored at $8$ bits. This is close to
full precision for these model sizes, but it is not identical, and we cannot
rule out that quantisation interacts with the methods differently.

\paragraph{Task domain, and testing methods outside their home ground.}
Both benchmarks are mathematics with automatically checkable answers. This
matters more than a usual scope caveat, because it is not neutral between the
methods. Self-Refine was proposed largely for open-ended generation, where an
answer can be better or worse in many small ways and a critique has something
to grasp. On a mathematics problem the answer is a single value that is either
right or wrong, so a critique has only one thing it can change and every change
it makes is a gamble. It is entirely possible that self-criticism helps on the
tasks it was designed for and hurts here.

We therefore do not claim that these methods do not work. We claim that on
mathematics with a checkable answer, at these model sizes, they lose to
spending the same tokens on more attempts. Anyone extending this to open-ended
tasks would need a different measure of quality than exact-match accuracy, and
that is a different experiment rather than a larger version of this one.

There is a further reason mathematics is the right place for \emph{this
particular} comparison, and it is worth stating because it cuts against us
rather than for us. Our baseline is majority voting, and majority voting needs
answers that can be compared for equality. On a summary or a piece of prose
there is no majority to take, so the baseline we are testing against does not
exist and the comparison cannot be run at all. Mathematics is therefore not the
domain where these methods look worst; it is the domain where their strongest
competitor is available. On open-ended work the practical alternative to
self-criticism is not majority voting but something weaker, such as picking one
sample arbitrarily, and against that weaker alternative self-criticism may well
win. Our result should be read as bounded by the availability of the baseline,
not as a claim that critique is useless wherever it is used.

\paragraph{One implementation per method.} Each method has many variants and
is sensitive to prompt wording. We implemented one faithful version of each
and used the same wording across models and datasets. A different
implementation could perform differently, and we release the exact prompts so
that this can be checked rather than argued about.

\paragraph{Fixed hyper-parameters.} We fixed the number of rounds, agents,
and samples in advance rather than tuning them per method. Tuning would have
raised every method's accuracy somewhat, but it would also have required
using the correct answers to choose the settings, which is precisely the leak
that \citet{huang2024selfcorrect} identify.

\subsection{A suggestion for method papers}

The protocol here is cheap to adopt. Reporting one extra curve, accuracy
against generated tokens for plain repeated sampling, on the same questions
and the same model, costs one pool of samples per benchmark and settles
the question of whether a method beats its own budget. We would encourage
authors of future test-time methods to include it, and reviewers to ask
for it.

\section{Conclusion}
\label{sec:conclusion}

Most methods that make a language model ``think harder'' also make it think
\emph{longer}, and longer buys accuracy by itself. Comparing such a method
against a single chain of thought therefore cannot show that its mechanism is
responsible for the gain. \citet{wang2024tokeneconomies} made this argument and
concluded that a simple sampling baseline frequently wins once budgets are
comparable. We asked whether that survives being tested rather than asserted,
and what explains it.

It survives, with a sharper shape than a blanket claim about elaborate
methods. In none of our four settings does any method significantly beat
repeated sampling at its own measured cost. But the split is not between
simple and elaborate, or between cheap and expensive. It is between methods
that aggregate independent attempts by counting and methods that ask the model
to evaluate its own work. Every method of the second kind, namely Self-Refine,
our forced version of Reflexion, and Best-of-$N$ with self-verification, is
below the equal-cost baseline in every comparison we ran. Methods that add no
self-assessment sit on the baseline rather than beneath it. Reflexion as its
authors define it is the exception that proves the point, since on the smaller
model it scored well only by declining to run.

Best-of-$N$ shows this without any confound. Given eight sampled solutions,
letting the model choose one is worse than counting which answer appears most
often, by $5$ to $17$ percentage points, on identical samples at identical
cost. The extra tokens are not the problem; what they are spent on is. A model
good enough to produce a correct answer among eight attempts is not
necessarily good enough to recognise it, and a tally does not have to
recognise anything.

Two smaller lessons seem worth carrying forward. First, a method whose control
flow is self-triggered can stop running without saying so: our Reflexion runs
scored well on the smaller model precisely because its self-assessment never
fired, making it a single chain of thought under another name. Evaluations of
adaptive methods should report how often the adaptive part engages. Second, an
analysis that infers what a method \emph{is} from the shape of what it stored
can silently score the wrong method. We did exactly that to Best-of-$N$ and
caught it only by cross-checking one number against another.

None of this settles what happens above $7$B, on open-ended tasks, or under a
cost measure that charges for input tokens, though on that last point the
omission runs against the methods we tested rather than in their favour.
What it does establish is that the comparison usually reported cannot support
the conclusions usually drawn from it, and that the missing comparison is
cheap: one pool of samples per benchmark yields the whole baseline curve. We
release the harness, the prompts, and all $19{,}200$ generations so the check
can be run rather than argued about.

\clearpage

\bibliographystyle{plainnat}
\bibliography{refs}

\clearpage
\appendix
\section{Reproducibility details}
\label{app:repro}

\subsection{Hardware and software}

All experiments ran on a single CloudLab machine: one AMD EPYC 7452
(32 physical cores, 64 hardware threads, Zen 2, AVX2 without AVX-512),
125\,GB RAM, Ubuntu 22.04, no GPU. Inference used \texttt{llama.cpp}
\citep{llamacpp} built from source with native optimisations, served through
\texttt{llama-server}. Most cells ran with $32$ parallel slots over a
$131{,}072$-token pool; the Best-of-$N$ cells were re-run with $16$ slots over
a $262{,}144$-token pool for the reason given in Appendix~\ref{app:repro}
below. Server configuration affects throughput and the maximum prompt that
fits, not the content of any generation.

Measured single-stream generation throughput was $57.8$ tokens/s for the
$1.5$B model and $31.1$ tokens/s for the $3$B model, both saturating at about
$16$ threads (the workload is memory-bandwidth bound). Batched serving raised
aggregate throughput to roughly $300$ tokens/s at $32$ concurrent requests, a
$5.2\times$ improvement, which is what made the study feasible on CPU.

\subsection{Prompts}

The system prompt is identical for all methods except the verification step
of Best-of-$N$:

\begin{quote}\footnotesize\ttfamily\raggedright
You are a careful problem solver. Reason step by step, then give the final
answer on the last line in the form \textbackslash boxed\{ANSWER\}.
\end{quote}

The base user prompt is the question followed by:

\begin{quote}\footnotesize\ttfamily\raggedright
Solve this. End with the final answer as \textbackslash boxed\{ANSWER\}.
\end{quote}

Method-specific prompts are:

\begin{description}\footnotesize\raggedright
\item[Plan-and-Solve] ``Let's first understand the problem and devise a plan
  to solve it. Then carry out the plan step by step.''
\item[Self-Refine, critique] ``Review your solution above. Point out any
  errors in the reasoning or arithmetic. Be specific and brief.''
\item[Self-Refine, revise] ``Feedback on your solution: \{feedback\}. Write an
  improved solution.''
\item[Reflexion, judge] ``Is your final answer correct? Reply with exactly
  CORRECT or INCORRECT, then one sentence of reasoning.''
\item[Reflexion, reflect] ``Your answer may be wrong. Write a short
  reflection on what specifically might have gone wrong.''
\item[Reflexion, retry] ``\{reflections\}. Using these reflections, solve the
  problem again.''
\item[Debate] ``\{other agents' solutions\}. Using these other answers as
  additional information, give your own updated solution.''
\item[Best-of-$N$, verify] ``\{solutions\}. Which solution is most likely
  correct? Reply with exactly: \texttt{BEST=}$\langle$number$\rangle$.''
\end{description}

\subsection{Seeding}

For each (model, dataset, method, configuration, question index) tuple we
compute a SHA-256 hash and take its first $32$ bits as the base seed. Within
a method, sub-calls derive their seeds from that base by fixed offsets. The
question subsample is drawn by shuffling the full test set with Python's
\texttt{random.Random(1234)} and taking the first $150$ items. Every run is
therefore reproducible exactly, and reruns resume from partial output by
skipping question indices already present in the output file.

\subsection{A failure that would have biased the results}

We record this because it is the kind of problem that is easy to absorb
silently. The server was first configured with $32$ parallel slots sharing a
$131{,}072$-token pool, giving each request $4{,}096$ tokens of context. That
is ample for every method except Best-of-$N$, whose verification step places
all eight candidate solutions into a single prompt. On MATH-500, where
solutions are long, that prompt exceeded the per-slot context and the server
returned HTTP 400. In total $151$ requests failed this way.

The damage was not the missing data but its \emph{pattern}. Requests failed
exactly when the eight solutions were long, and solution length rises with
problem difficulty, so the surviving records were a biased sample of easier
questions. The largest total generation among surviving Best-of-$N$ runs
was $3{,}997$ tokens, just under the $4{,}096$ limit, which is the signature
of a truncation rather than a random fault. Analysing what survived would
have credited Best-of-$N$ with an accuracy measured on an easier subset of
problems than every other method was given.

We therefore re-ran the affected cells with $16$ slots over a
$262{,}144$-token pool ($16{,}384$ tokens per slot) until every method had all
$150$ questions. Records that had already succeeded were reused unchanged:
they were correct results, and because the runner skips question indices it
has already written, resuming adds only the missing questions. The final
dataset has no failed requests.

The size of the distortion is worth stating, because it shows this was not a
hypothetical concern. On Qwen2.5-1.5B with MATH-500, chain-of-thought scores
$67.9\%$ on the $78$ questions that survived truncation and $47.3\%$ on the
full $150$; Best-of-$N$ scores $75.6\%$ against $58.0\%$. The truncated sample
was roughly twenty percentage points easier. The conclusions moved as well:
forced Reflexion goes from $-11.1$ points against the cost-matched baseline
with a Holm-corrected $p$ of $0.143$ on the truncated sample, to $-8.9$ points
with $p = 0.043$ on the complete one, a smaller effect that is nonetheless
significant, because the full sample is both larger and harder. After the
refill, the longest surviving Best-of-$N$ run generates $7{,}192$ tokens,
comfortably above the old ceiling, confirming that the previously missing
records were the long ones.

\subsection{A scoring bug we found by cross-checking, and what it changed}

Our analysis re-grades every record from the stored raw text so that the
runner and the analysis cannot disagree. That code branched on what a record
contained: if it held a list of generations, the answer was taken to be the
majority vote over them. That is right for debate, whose agents are aggregated
by vote, and wrong for Best-of-$N$, whose defining step is that the
\emph{model} selects one candidate. Best-of-$N$ stores its candidates as such a
list, so it was silently scored as though it were self-consistency, and the
verifier's choice, which is the entire method, was discarded.

Nothing in the results looked anomalous. We found it only by computing the
verifier's accuracy separately, for the mechanism analysis in
Section~\ref{sec:results}, and noticing that the main table's Best-of-$N$
figure matched the majority-vote number to the decimal place in all four
settings rather than the verifier's, which was $8$ to $17$ points lower.

The correction matters. Under the bug, Best-of-$N$ appeared to sit on the
baseline ($+0.3$ to $+3.2$ points, never significant). Scored correctly, it is
below the baseline in every setting, significantly so in three, and it
supplies the paper's only result that survives Holm correction across all
$28$ comparisons at once. We report this because a reader is entitled to know
that the headline number changed after a bug fix, and because the failure mode
generalises: an analysis that infers what a method \emph{is} from the shape of
what it stored will quietly evaluate the wrong method.

\subsection{An independent re-derivation of every number}

We found two scoring problems in this project, described above. Neither was
visible in the results, and both were caught by comparing one number against
another rather than by any systematic check. That is not a comfortable basis
on which to ask a reader to trust the rest, so we wrote a second program that
recomputes the paper's quantities from the stored records without importing
the analysis code, using its own aggregation.

It re-grades every method from raw text, recomputes each method's accuracy and
mean token cost, recomputes the Best-of-$N$ verifier-versus-majority gap,
checks that no result file contains a duplicate or missing question, and
checks that the per-sample token counts in each baseline pool sum to the total
recorded for that question. Across the four settings this is $96$ independent
checks, and all $96$ agree with the numbers reported here.

Agreement between two implementations is not proof that both are right, since
they share the grading module and the same underlying records. It does rule
out the class of error that actually bit us twice, namely an aggregation step
that quietly computes something other than what it claims. The script is
included in the release so the check can be repeated rather than believed.

\subsection{Cost accounting}

Token counts are taken from the server's own \texttt{usage.completion\_tokens}
field for every call, and summed across all calls a method makes for a
question. No count is estimated or inferred. For the sampling baseline we
additionally record per-sample token counts, which is what allows the cost of
a subsampled self-consistency run to be computed exactly rather than
approximated by an average.

\end{document}